\documentclass[letterpaper, 10 pt, conference]{ieeeconf}  

\IEEEoverridecommandlockouts                              

\usepackage{amsfonts}
\usepackage{cite}
\usepackage{amsmath}
\usepackage{booktabs}
\usepackage{bm}
\usepackage{multirow}
\usepackage{tabularx}
\usepackage{graphicx}
\usepackage{subcaption}
\usepackage[table,xcdraw]{xcolor}
\usepackage{caption}
\usepackage{marvosym}
\usepackage{graphicx}
\usepackage{hyperref}

\title{\LARGE \bf
SceneDM: Scene-level Multi-agent Trajectory Generation with Consistent Diffusion Models
}

\author{Zhiming Guo\textsuperscript{*}, Xing Gao\textsuperscript{*, \Letter} , Jianlan Zhou, Xinyu Cai, and Botian Shi% 
\thanks{* Equal contribution.}%
\thanks{Zhiming Guo is with School of Artificial Intelligence and Automation, Huazhong University of Science and Technology, and Shanghai Artificial Intelligence Laboratory.
}%
\thanks{Jianlan Zhou is with School of Artificial Intelligence and Automation, Huazhong University of Science and Technology. 
}%
\thanks{Xing Gao, Xinyu Cai, and Botian Shi are with Shanghai Artificial Intelligence Laboratory. 
}%
\thanks{\Letter \ Corresponding to Xing Gao,  gxyssy@163.com.}
}
\begin{document}

\maketitle
\thispagestyle{empty}
\pagestyle{empty}

\begin{abstract}
Realistic scene-level multi-agent motion simulations are crucial for developing and evaluating self-driving algorithms. However, most existing works focus on generating trajectories for a certain single agent type, and typically ignore the consistency of generated trajectories. In this paper, we propose a novel framework based on diffusion models, called SceneDM, to generate joint and consistent future motions of all the agents, including vehicles, bicycles, pedestrians, etc., in a scene. To enhance the consistency of the generated trajectories, we resort to a new Transformer-based network to effectively handle agent-agent interactions in the inverse process of motion diffusion. In consideration of the smoothness of agent trajectories, we further design a simple yet effective consistent diffusion approach, to improve the model in exploiting short-term temporal dependencies. Furthermore, a scene-level scoring function is attached to evaluate the safety and road-adherence of the generated agent's motions and help filter out unrealistic simulations. Finally, SceneDM achieves state-of-the-art results on the Waymo Sim Agents Benchmark. Project webpage is available at \href{https://alperen-hub.github.io/SceneDM}{https://alperen-hub.github.io/SceneDM}.
\end{abstract}

\section{INTRODUCTION}
Traffic simulations complement real-world logged traffic scenarios, providing an economical and safe way to evaluate autonomous driving systems before their deployment in the real world. However, generation of such scenes is nontrivial, because of (i) diverse agent types, including vehicles, pedestrians, bicycles, etc.,
and their complex  interactions; (ii) multi-modal nature of the generated scenes.

Several rule-based strategies \cite{dosovitskiy2017carla,hay2005sumo} provide some intuitive solutions but struggle to provide complex traffic scenes. Alternatively,  recent works resort to deep models to handle the complexity of traffic scenes. For example, some works \cite{feng2023trafficgen,tan2021scenegen,varadarajan2022multipath++} exploit motion prediction methods to obtain future trajectories of agents.
On the other hand, generative models are exploited, including Generative Adversarial Nets (GAN) based methods
\cite{haakansson2021driving,bhattacharyya2022modeling,ding2021multimodal,yin2021diverse} and
Variational Auto-encoder (VAE) ones \cite{ding2019multi,oh2022cvae}. 
Besides, a variety of diffusion-based methods have been proposed recently,
such as MID\cite{gu2022stochastic}, MotionDiffuser\cite{jiang2023motiondiffuser}, and CTG\cite{zhong2023guided}.

\begin{figure}
\centering
\includegraphics[width=1.0\linewidth]{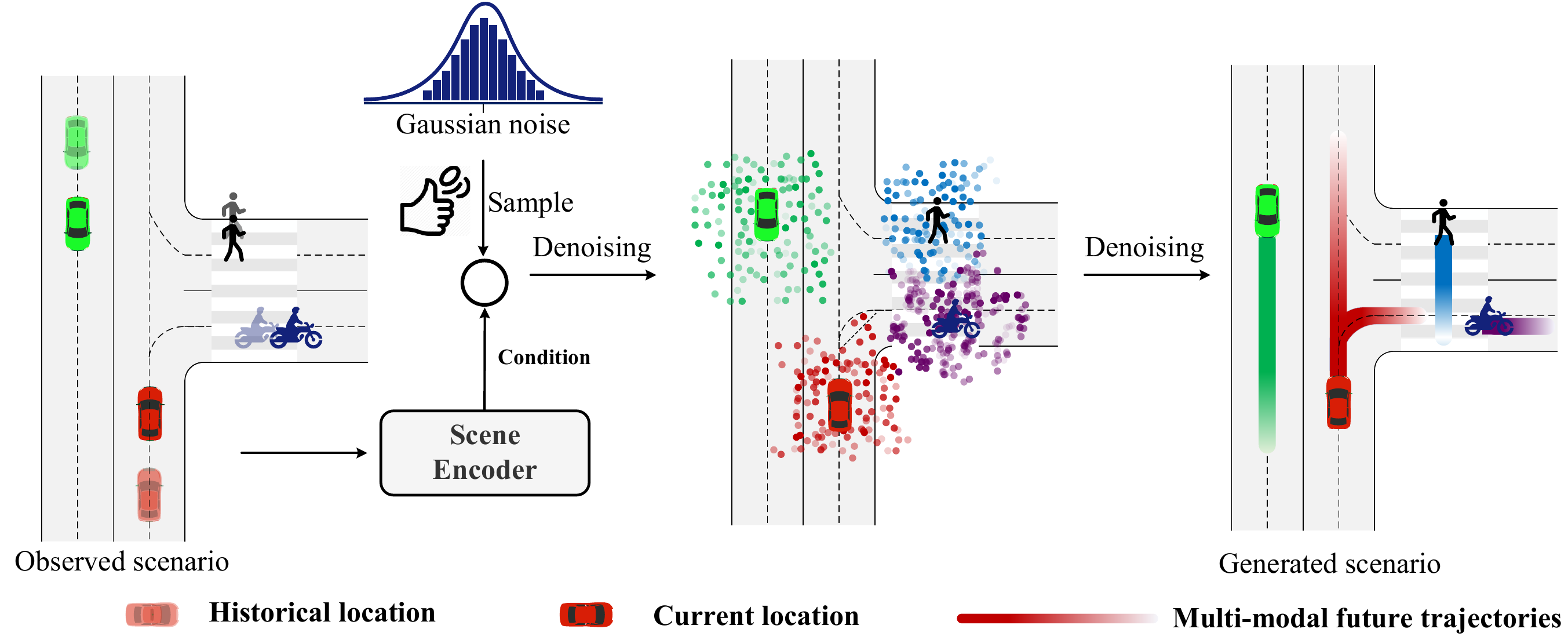} 
\caption{ Conditioned on road networks and historical trajectories of agents, SceneDM as a diffusion model generates diverse traffic scenes composed of consistent future motions of all the observed agents, including vehicles, pedestrians, and bicycles, from Gaussian noise.}\label{fig:issue}
\end{figure}

However, these methods still have some limitations.
First, existing methods usually focus on generating future trajectories for a single type of agent  
such as pedestrians \cite{gu2022stochastic} or vehicles \cite{zhong2023guided}, rather than all the types of agents in a scene. Furthermore, most existing generative methods fail to exploit the local smoothness of the trajectories of agents, which will lead to unrealistic results.

To tackle these problems, we propose a novel diffusion framework, termed SceneDM,  for scene-level multi-agent trajectory generation. SceneDM generates realistic future trajectories for multiple types of agents based on historical trajectories and map information. The proposed diffusion strategy effectively enhances the local smoothness of the generated trajectories and further improves on the consistency of the future motion  of different agents in a scene. 

For locally smooth agent trajectories like in the real world, we introduce consistent diffusion for the trajectory generation, given the observation in image domain that the noise has a large impact on the pattern of the generated image \cite{chen2023control}. Generally, in order to  improve on the consistency of the trajectory, we attempt to make adjacent motion states of the trajectory to have partly overlapped noise in the diffusion process. This is non-trivial, however, for simply imposing the same noise to all states will result in constant sequence data. To this end, we propose a scheme to augment the sequence of motion states by concatenating each state with its following  state.  In this way, adjacent states of the augmented trajectory have overlapping parts, and we apply the same noise to this overlapping part during the diffusion process to improve the similarity of the generated adjacent motion of the augmented trajectory and thereby enhance the local smoothness of the original trajectory. Correspondingly, a temporal-consistent guidance sampling strategy is proposed to sample trajectories once the diffusion model has been trained with the proposed strategy. Furthermore, through trajectory augmentation, the difference between the front and rear halves of each augmented state corresponds to the change in the adjacent states of the original trajectory. With a designed regularization of the difference,  we further improve the smoothness of the generated data.

Furthermore, to capture the dependencies of future motions of agents, we adopt temporal and spatial attention modules for the denoising network.  It is a Transformer-based network  that consists self-attention module across time and agent alternately. Finally, a scene-level scoring function is presented to evaluate 
 the traffic regulation compliance of the generated scenarios from the perspectives of collision and road-adherence. This scoring function help filter out unrealistic simulations, ensuring the practicality of the generated samples. The main contributions of SceneDM are summarized as follows: 

\begin{itemize}
\item We design a novel diffusion model based framework, termed SceneDM, that jointly generates future trajectories for various types of agents while maintaining scene consistency.  SceneDM achieves state-of-the-art results on the public Waymo Open Sim Agents Benchmark.
\item We propose a consistent diffusion strategy to improve the correlation between adjacent motion states of generated trajectory  and produce locally smooth agent trajectories.
\item We introduce a scene-level scoring function to 
select the generated traffic scenes that comply with traffic regulations. 
\end{itemize}

\section{RELATED WORK}
We provide an overview of a series of traffic simulation methods, including trajectory prediction induced methods and generative model based works.

\textbf{Motion prediction induced methods.} Some recent works exploit the results of motion prediction tasks to generate multi-modal traffic scenes. For instance, Simnet \cite{bergamini2021simnet} models the vehicle's driving process as a Markov process and implements state distribution and transition function with deep neural networks.  TrafficSim \cite{suo2021trafficsim} formulates a joint actor policy with an implicit latent variable model and employs GRU and CNN to learn multi-agent behaviors from real-world data. Besides, Trafficgen \cite{feng2023trafficgen}, derived from the  motion forecasting model Multipath++ \cite{varadarajan2022multipath++},  utilizes multi-context gating blocks to handle various interactions in observed data and exploits Gaussian mixture models to model the multi-modal characteristics of the generated trajectories. Furthermore, MTR+++ \cite{qian20232nd} introduces a  collision-mitigation policy to improve the trajectories produced from the motion prediction model MTR \cite{shi2022motion}.

\textbf{Generative models.} Another category of approaches utilizes generative models to learn the probability distribution of trajectory data and generate new trajectory samples.
Some methods \cite{ding2021multimodal,yin2021diverse} utilize GAN to generate multiple trajectories for traffic agents.
However, GANs may suffer from mode collapse and will produce unrealistic scenarios \cite{jiao2022tae}.
In addition, MTG \cite{ding2019multi} and CVAE-H \cite{oh2022cvae} employ VAE to extract  representations of  historical trajectories of agents and generate future trajectories.
In a more relevant work, MID \cite{gu2022stochastic} presents a diffusion-based framework to formulate pedestrian trajectory prediction. Besides,
Scene Diffusion \cite{pronovost2023generating} utilizes latent diffusion in an end-to-end differentiable architecture to 
generate arrangements of discrete bounding boxes for agents.
CTG \cite{zhong2023guided} develops a conditional diffusion model for controlled vehicle trajectory generation, 
ensuring that the generated  trajectories with  desired properties, such as speed limits. However, these diffusion-based models only consider trajectory generation for a single type of agent.

In contrast with these methods, we propose a consistent diffusion framework to generate trajectories with scene consistency for various types of agents in the scene, 
including vehicles, pedestrians, and bicycles. 
Additionally, we design a novel diffusion strategy to address the temporal consistency of generated trajectories 
and improve the local smoothness of the trajectories.
\begin{figure*}
\centering
\includegraphics[width=1.0\linewidth]{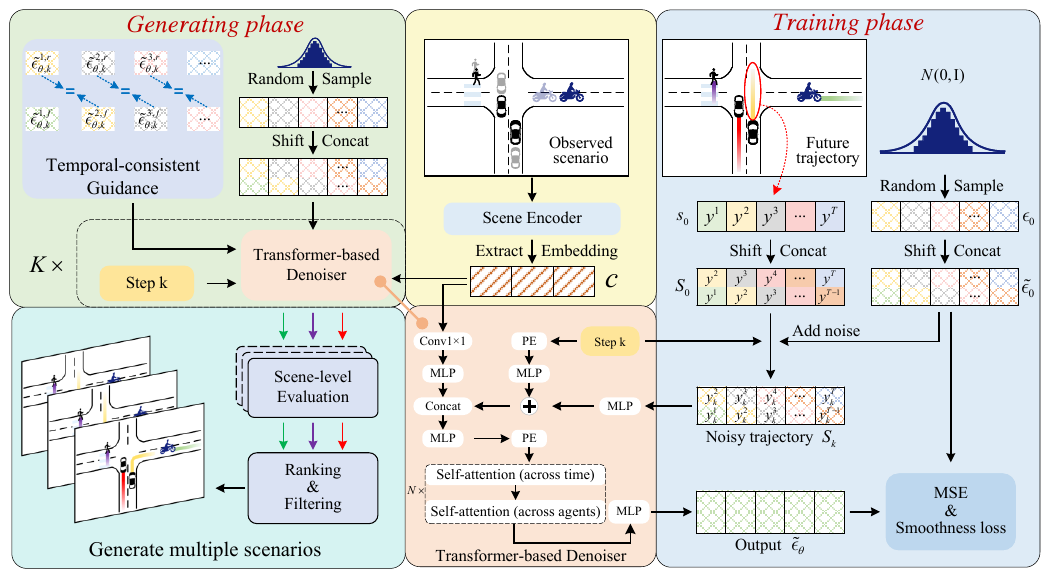} 
\caption{The framework of SceneDM. SceneDM consists of a scene encoder and a Transformer-based denoiser network. The denoiser network utilizes the agent embedding learned by the scene encoder to remove noise from the noisy trajectory. During training, SceneDM first augments the  trajectory sequence  by concatenating adjacent motion states and imposes the same noise to the overlapping part. Models are then optimized to predict the added noise. Furthermore, we introduce a smoothness regularization to improve the generated trajectory's smoothness. At the generating phase, we achieve temporal consistency of the latent variable  $\bm{S_k}$ through the proposed temporal-consistent  guidance. Finally, a scene-level scoring module is designed to filter out unrealistic simulations, ensuring the practicality of the generated samples.} \label{fig:main}
\end{figure*}

\section{METHOD}
\subsection{Notions and Preliminaries}
SceneDM aims to generate future trajectories for $N$ agents in a given scenario simultaneously, 
leveraging both the map information and their historical trajectory data.
In this paper, the current time is denoted as $t=0$. 
The future trajectory of an agent is represented as 
$\bm{s}={\left\{\bm{y^{t}} \in \mathbb{R}^{H}\right.}|t=1,2,\cdot\cdot\cdot,T\mathbf{\rbrace}$, where $\bm{y^{t}}$ is an $H$-dimensional vector including 3-D coordinates and heading and $T$ represents the length of the generated future trajectory. 

Diffusion models consist of a diffusion process that gradually transforms a data distribution into unstructured noise 
and a reverse process to recover the data distribution \cite{sohl2015deep,ho2020denoising}. In this paper, we employ subscripts to indicate the step in the diffusion process and reverse process, such as original data $\bm{s_0}$ and latent variable $\bm{s_k}$.
During the forward diffusion process, Gaussian noise is gradually added to the original data $\bm{s}_0$ to obtain latent variable $\bm{s}_1, \cdots, \bm{s}_K$,
where $K$ denotes the maximum number of diffusion steps.
According to DDPM \cite{ho2020denoising}, the forward diffusion process is parameterized as a Markov chain, with the final variable,
\begin{equation}
    \bm{s}_K=\sqrt{\bar{\alpha}_{K}}\bm{s}_0 +\sqrt{(1-\bar{\alpha}_{K})}\bm{\epsilon},
\end{equation}
where $\bar{\alpha}_{K}$ is a positive constant
representing the noise level and $\bm{\epsilon}$ denotes the noise sampled from the Gaussian distribution ${\cal N}(0,\mathrm{I})$. When $K$ is sufficiently large, $\bm{s}_K$ converges to the Gaussian distribution.

Diffusion models provide parameterized Gaussian transitions to model the reverse process. 
Given diffusion step $1, \cdots, K$ and the condition $\bm{c}$, diffusion models formulate the reverse process as follows:
\begin{gather}
    p_{\theta}\left(\bm{s}_{0:K} \mid \bm{c}\right) =p\left(\bm{s}_{K} \mid \bm{c}\right) \prod_{k=1}^{K} p_{\theta}\left(\bm{s}_{k-1} \mid \bm{s}_{k}, \bm{c}, k\right), \\
    p_{\theta}\left(\bm{s}_{k-1} \mid \bm{s}_{k}, \bm{c}, k\right) =\mathcal{N}\left(\boldsymbol{\mu}_{\theta}\left(\bm{s}_{k}, \bm{c}, k\right), \mathbf{\Sigma}_{\theta}\left(\bm{s}_{k}, \bm{c}, k\right)\right),
\end{gather}
where $p\left( \bm{s}_{K} \mid \bm{c} \right) = p\left(\bm{s}_{K} \right) = {\cal N}( \bm{0},\bm{I})$
and the $\theta$ indicates the parameters of the entire framework.

Diffusion models \cite{ho2020denoising} are optimized to approximate $p_{\theta}\left(\bm{s}_{k-1} \mid \bm{s}_{k}, \bm{c}, k\right)$ or 
equivalently predict the added noise $\bm{\epsilon}$ in the diffusion process, in accordance with the objective function
\begin{equation}
    L_{mse}=\mathbb{E}_{\bm{\epsilon}, \bm{s}_{0}}\left\| \bm{\epsilon}-\epsilon_{\theta}(\bm{s}_{k}, k, \bm{c})\right\|,
\end{equation}
with $\epsilon_{\theta}(\bm{s}_{k}, k, \bm{c})$ representing the predicted noise.

\subsection{Framework}
The framework of SceneDM consists of a scene encoder to learn vectorized representations of dynamic scenarios and a designed decoder for the reverse diffusion process, 
as illustrated in Fig.~\ref{fig:main}. 
The scenario encoder, a scene-centered model \cite{gao2023dynamic} is adopted as an example, 
encodes scene elements such as road networks and agent historical trajectories into a set of latent embeddings. 
The embedding of agents are then fed into the Transformer-based decoder as the condition in the reverse process.

In the design of decoder, we adopt attention mechanism to handle multi-agent interactions and temporal dependencies of trajectories.  The temporal attention enables the model to learn continuous trajectories over time, irrespective of the agent's identity. 
Meanwhile, the spatial attention  permits the model to capture agent-agent interactions and generate consistent trajectories, \emph{i.e.,} without collision. Specifically, we take alternating temporal attention layer and spatial attention layer as a basic module, similarly to \cite{ngiam2021scene}. Multiple such modules are stacked to process the complex interactions in the denoising process. 

As illustrated at the bottom of Fig.~\ref{fig:main}, with the embedding $\bm{c}$ from the encoder, for each diffusion step $k$, we first encode the step and noisy variable $\bm{s_k}$ through multi-layer perceptrons (MLPs). These embeddings are further combined with the condition $\bm{c}$ into a fused feature. To highlight the positional relationships of the sequence data $\bm{s}_0=[\bm{y^1}, \bm{y^2}, \cdots, \bm{y^T}]$,  we further impose positional encoding  on the fused feature. This feature is then fed into the transformer that is composed of attention layers across time and agent alternatively. Finally, an MLP based on the features from the transformer produces the noise to be removed.

\subsection{Consistent Diffusion}
In real-world scenarios, agents exhibit smooth and continuous motion patterns as they navigate through their environment.
Therefore, local smoothness is an important factor in evaluating the reality of generated trajectories. To address it, we propose a novel consistent diffusion approach to generate smooth and realistic trajectories through imposing  the same noise on the overlapping parts of adjacent elements.

\textbf{Diffusion process.} As shown in Figure \ref{fig:main}, we firstly augment the trajectory sequence $\bm{s_0} = [\bm{y^1}, \bm{y^2}, \cdots, \bm{y^T}]$ to make adjacent elements have overlapping parts. Specifically, for each state $\bm{y^t}$ in the trajectory $\bm{s_0} = [\bm{y^1}, \bm{y^2}, \cdots, \bm{y^T}]$, we concatenate it with the state $\bm{y^{t+1}}$ at the subsequent moment.
By concatenating the states of adjacent frames, we create an augmented variable $\bm{\tilde{y}^t}\in \mathbb{R}^{2H}$ that incorporates information from both $\bm{y^t}$ and $\bm{y^{t+1}}$, 
providing an informative input for subsequent processing. 
Most importantly, $\bm{\tilde{y}^t}$ overlaps partly with both $\bm{\tilde{y}^{t-1}}$ and $\bm{\tilde{y}^{t+1}}$.
Correspondingly, we obtain an augmented sequence of trajectory states, denoted as $\bm{S_{0}}= [\bm{\tilde{y}^1}, \bm{\tilde{y}^2}, \cdots, \bm{\tilde{y}^{T-1}} ]$. We then gradually add Gaussian noise to $\bm{S_{0}}$ to obtain $\bm{S_{k}}$. To maintain the consistency of the state information at the same timestamp within the noisy trajectory $\bm{S_{k}}$, we apply the same augmentation approach to the sampled noise. 
Specifically, we first sample noise $\bm{\epsilon_{0}} = [\bm{\epsilon_{0}^1}, \bm{\epsilon_{0}^2}, \cdots, \bm{\epsilon_{0}^T}]^T \in \mathbb{R}^{T\times H}$ from ${\cal N}(0,\mathrm{I})$. 
Then $\bm{ \epsilon_{0}}$ is shifted and concatenated to obtain the augmented noise sequence $\bm{\tilde{\epsilon}_0} \in \mathbb{R}^{T-1\times 2H}$ for $\bm{S_0}$. In other words,  
\begin{gather}
\bm{\tilde{\epsilon}_{0}^t} = {{\rm Concat}( \bm{\epsilon_0^t}, \bm{\epsilon_0^{t+1}}}), t=1, 2, \dots, T-1,\\
\bm{S_K}=\sqrt{\bar{\alpha}_{K}}\bm{S_0} +\sqrt{(1-\bar{\alpha}_{K})}\bm{\tilde{\epsilon}_0}.
\end{gather}

\textbf{Reverse process.} Together with the condition $\bm{c}$ from encoder, $\bm{S_{k}}$ is passed through the Transformer denoiser to predict the noise $\tilde{\epsilon}_{\theta}(\bm{S_{k}}, k, \bm{c}) \in \mathbb{R}^{T-1 \times 2H}$ to be removed.
The model is optimized as:
\begin{equation}
    L_{mse}=\mathbb{E}_{\bm{\epsilon}, \bm{S}_{0}}\left\|\bm{\tilde{\epsilon}_{0}}-\tilde{\epsilon}_{\theta}(\bm{S_{k}}, k, \bm{c})\right\|, \ \ k=1, 2, \dots, K.
\end{equation}

Besides,  we introduce a regularization to further improve smoothness. The smoothness is  calculated as the difference between adjacent motion states of the original sequence data, and equivalently the difference between the front and rear halves of the element of the augmented sequence. For example, with motion state indicating linear velocity of the agent, the difference reflects the linear acceleration.
The smoothness loss term regularizes the difference of the generated samples to approximate to that of the ground truth that is observed in the real-world.  Mathematically, 
\begin{equation}
L_{smooth}=\mathbb{E}_{\bm{\epsilon}, \bm{S_{0}}, k, t}\left\|(\bm{\epsilon_{0}^{t+1}}-\bm{\epsilon_{0}^t})- (\bm{\tilde{\epsilon}_{\theta, k}^{t,r}}-\bm{\tilde{\epsilon}_{\theta, k}^{t,f}})\right\|,
\end{equation}
where we use $\bm{\tilde{\epsilon}_{\theta, k}^{t,f}}$  and $\bm{\tilde{\epsilon}_{\theta, k}^{t,r}}$  to indicate the first and rear part of each predicted noise $\bm{\tilde{\epsilon}_{\theta, k}^t}$, respectively.

By combining the aforementioned two losses, we define a novel hybrid optimization objective:
\begin{equation}
    L_{hybrid}=L_{mse}+\lambda L_{smooth}. \label{eq:hybrid_loss}
\end{equation} 
By introducing the smoothness regularization term in the overall loss function, 
the model is encouraged to generate trajectories that exhibit realistic and smooth motion patterns. 
Both parameters of the scene encoder and the Transformer-based denoiser are trained simultaneously.
The hyperparameter $\lambda$ is used to adjust the balance between these two losses to prevent $L_{smooth}$ from
overwhelming $L_{mse}$.

\begin{table*}[t]
    \caption{Results on the public Waymo Sim Agents Benchmark. Realism meta-metric is the primary metric for ranking the methods and  higher value indicate better model performance. The best two results are indicated in \textbf{bold} and \underline{underlined}.}
    \label{tab:leader_result}
    \begin{tabular}{@{}lllllllllll@{}}
    \toprule
    & \multicolumn{1}{c}{\textbf{Meta Metric}} & \multicolumn{4}{c}{Kinematic Metric}  & \multicolumn{3}{c}{Interactive Metric}   & \multicolumn{2}{c}{Map Metric}    \\ \cmidrule(l){2-11} 
    & \textbf{Realism}   & \begin{tabular}[c]{@{}l@{}}Linear\\ Speed\end{tabular} & \begin{tabular}[c]{@{}l@{}}Linear\\ Acceleration\end{tabular} & \begin{tabular}[c]{@{}l@{}}Angel\\ Speed\end{tabular} & \begin{tabular}[c]{@{}l@{}}Angel\\ Acceleration\end{tabular} & \begin{tabular}[c]{@{}l@{}}Dist To\\ Object\end{tabular} & Collision & \begin{tabular}[c]{@{}l@{}}Time To\\ Collision\end{tabular} & \begin{tabular}[c]{@{}l@{}}Dist To\\ Roadedge\end{tabular} & Offroad \\ \midrule
    SceneDMF      & \textbf{0.5060}  &  0.4315       & \textbf{0.2767}  & 0.5230 & \underline{0.4666}  & \underline{0.3678}  & \textbf{0.4625}    & \textbf{0.8128} & 0.6215  & \textbf{0.5985} \\
    SceneDM  & \underline{0.5000}    & 0.4316      & \underline{0.2765}  & \underline{0.5232} & \underline{0.4666}  & 0.3650       & \underline{0.4473}    & 0.8102    & 0.6195  & \underline{0.5843}  \\
    \midrule
    MTR\_E  & 0.4911 & 0.4278 & 0.2353  & \textbf{0.5335}  & \textbf{0.4753} & 0.3455 & 0.4091    & 0.7983  &\underline{0.6541} & 0.5840  \\
    Multipath & 0.4888 & \underline{0.4318} & 0.2304 & 0.5149 & 0.4521  & 0.3440 & 0.4198    & \underline{0.8127} & 0.6394  & 0.5830  \\
    MTR+++   & 0.4697 & 0.4119 & 0.1066  & 0.4838  & 0.4365  & 0.3457 & 0.4144    & 0.7969  & \textbf{0.6545} & 0.5770  \\
    CAD     & 0.4321 & 0.3464 & 0.2526  & 0.4327  & 0.3110  & 0.3300 & 0.3114    & 0.7893  & 0.6376 & 0.5397  \\
    QCNeXt  & 0.3920 & \textbf{0.4773} & 0.2424  & 0.3252 & 0.1987   & \textbf{0.3759} & 0.3244    & 0.7569  & 0.6099 & 0.3600  \\
    sim\_agents\_tutorial  & 0.3201 & 0.3826 & 0.0999  & 0.0318 & 0.0391  & 0.2909  & 0.3360 & 0.7549    & 0.5251  & 0.3804  \\
    \bottomrule
    \end{tabular}
    \end{table*}

\subsection{Temporal-consistent Guidance Sampling}
We present here the temporal-consistent guidance sampling strategy for the consistent diffusion approach, generalized from DDIM \cite{song2020denoising}. As discussed before, SceneDM achieves local smoothness through sequence element augmentation and a tailored noise imposition strategy at the training stage. 
Correspondingly, during the generation phase, after sampling noise from a Gaussian distribution, 
we perform shifting and concatenation operations to the sampled noise sequence to obtain the initial noisy trajectory sequence $\bm{S_{K}}\in \mathbb{R}^{T-1 \times 2H}$.
Furthermore, we introduce temporal-consistent guidance to enhance the temporal consistency of the generated trajectories.  At each reverse step, the denoiser predicts the noise.
During the denoising process, 
temporal-consistent guidance ensures the information corresponding to the same state of the original sequence data to be consistent, 
like $\bm{\tilde{\epsilon}_{\theta, k}^{t-1,r}}$ and $\bm{\tilde{\epsilon}_{\theta, k}^{t,f}}$, 
through averaging them:
\begin{equation}
   \bm{\tilde{\epsilon}_{\theta, k}^{t-1,r}}, \bm{\tilde{\epsilon}_{\theta, k}^{t,f}} \leftarrow \text{Mean}(\bm{\tilde{\epsilon}_{\theta, k}^{t-1,r}}, \bm{\tilde{\epsilon}_{\theta, k}^{t,f}}), \ \ t=2, 3, \dots, T.
\end{equation}

During the sampling process, SceneDM refines and generates the trajectories by iteratively computing transitions from $k=K$ to $k=0$ as follows:
\begin{align}
\bm{S_{k-1}} = &\sqrt{\frac{\bar{\alpha}_{k-1}}{\bar{\alpha}_{k}}} \bm{S_{k}} + \sqrt{1-\bar{\alpha}_{k-1}} \cdot \bm{\tilde{\epsilon}_{\theta,k}} \nonumber \\
& - \sqrt{\frac{\bar{\alpha}_{k-1}(1-\bar{\alpha}_{k}) }{\bar{\alpha}_{k}}} \cdot \bm{\tilde{\epsilon}_{\theta,k}},
\end{align}
where $\bar{\alpha}_{k}$ represent the noise levels at diffusion step $k$. 
By iteratively applying the transitions, the sampling process
gradually removes the noise and generates plausible future trajectories.

\subsection{Scene-level Scoring Module}
Generative models may produce unrealistic scenes or  traffic-rule violation ones, such as agents colliding or going out of the road boundary.  Such data may adversely affect the subsequent simulation task for autonomous driving. To address this issue, we propose a scoring module to assess the generated scenes at a scene-level.  It consists of safety verification and road-adherence measurements.

Specifically, safety verification is performed by calculating the overlapping area between agent bounding boxes.
Referring to \cite{ganjugunte2007survey}, we calculate the penetration depth to determine the maximum overlap distance between any two agents.
A positive value indicates a collision.
For each generated candidate trajectory $\bm{s_{i}}$, we compute its overlap with the trajectories of other agents in the scene in parallel and denote the number of collisions  as $r_{1}(s_{i})$.
Similarly, the road-adherence is measured by $r_{2}(s_{i})$, which represents the number of times the agent goes out of the road boundary. The final trajectory scoring function for $\bm{s_i}$ is obtained by:
\begin{equation}
    F(s_{i})=r_{1}(s_{i}) + r_{2}(s_{i}).
\end{equation}
Notably, the lower $F(s_{i})$ is, the more the generated trajectory $s_{i}$ is in compliance with the traffic regulations.  Through calculating the average of   $F(\bm{s_{i}})$, $ i = 1,2,\cdots,N$, we obtain the score of the generated scenario.

\section{Experiment and Results}
We evaluate our model on the public Waymo Sim Agents Benchmark, where the proposed
model achieves state-of-the-art results. 
Several ablation studies are  further conducted to verify the proposed modules.

\subsection{Dataset and Metrics}
We use the Waymo Open Motion Dataset in our experiment. 
The dataset comprises 486,995, 44,097, and 44,920 scenarios in the training, validation, and test set, respectively. 
In each scenario, there are a maximum of 128 agents, consisting of three types: vehicles, bicycles, and pedestrians. 
The dataset provides the historical trajectory information of these agents for a duration of 1.1 seconds, 
including 3D coordinate (x, y, z), heading, vehicle speed, and the shape of the agent. 
Additionally, the dataset also provides the map information of the scene.

The Sim Agents Benchmark requires simulating 32 future trajectories for each agent in the scene, 
generating their motions for the upcoming 8-second duration. 
Detailed simulation information should include the centroid coordinates and heading.
The Benchmark evaluates the similarity between the distribution of the generated trajectories and the distribution of real-world data. 
This is accomplished by calculating the approximate negative log-likelihood.
The primary evaluation metric of the benchmark is the \textbf{realism} meta-metric, which incorporates a comprehensive set of component metrics, 
including kinematic, interactive, and map-based metrics. 
For more detailed information, please refer to \cite{montali2023waymo}.

\subsection{Implementation Details}
We utilize a scene-centric coordinate system where all agents within the scene share the same coordinate system. 
We take the location of the autonomous vehicle at $t=0$ as the origin, and adopt its current driving direction as x-axis. 
Instead of directly generating 3D coordinates of agents, we choose to generate agents' velocities and integrate them to generate trajectories.
Within the decoder, we perform six iterations of the attention mechanism, alternating between the time dimension and the agent dimension.
The embedding dimension in the decoder is set to 512.
During the training phase, the initial learning rate is set to 0.0001 and decreases with the step decay learning rate scheme. We set the loss weight $\lambda$ in Equation \ref{eq:hybrid_loss} to 1.0.

\begin{figure*}[htbp]
    \centering
    \begin{subfigure}{5.85cm}
        \centering
        \includegraphics[width=\textwidth]{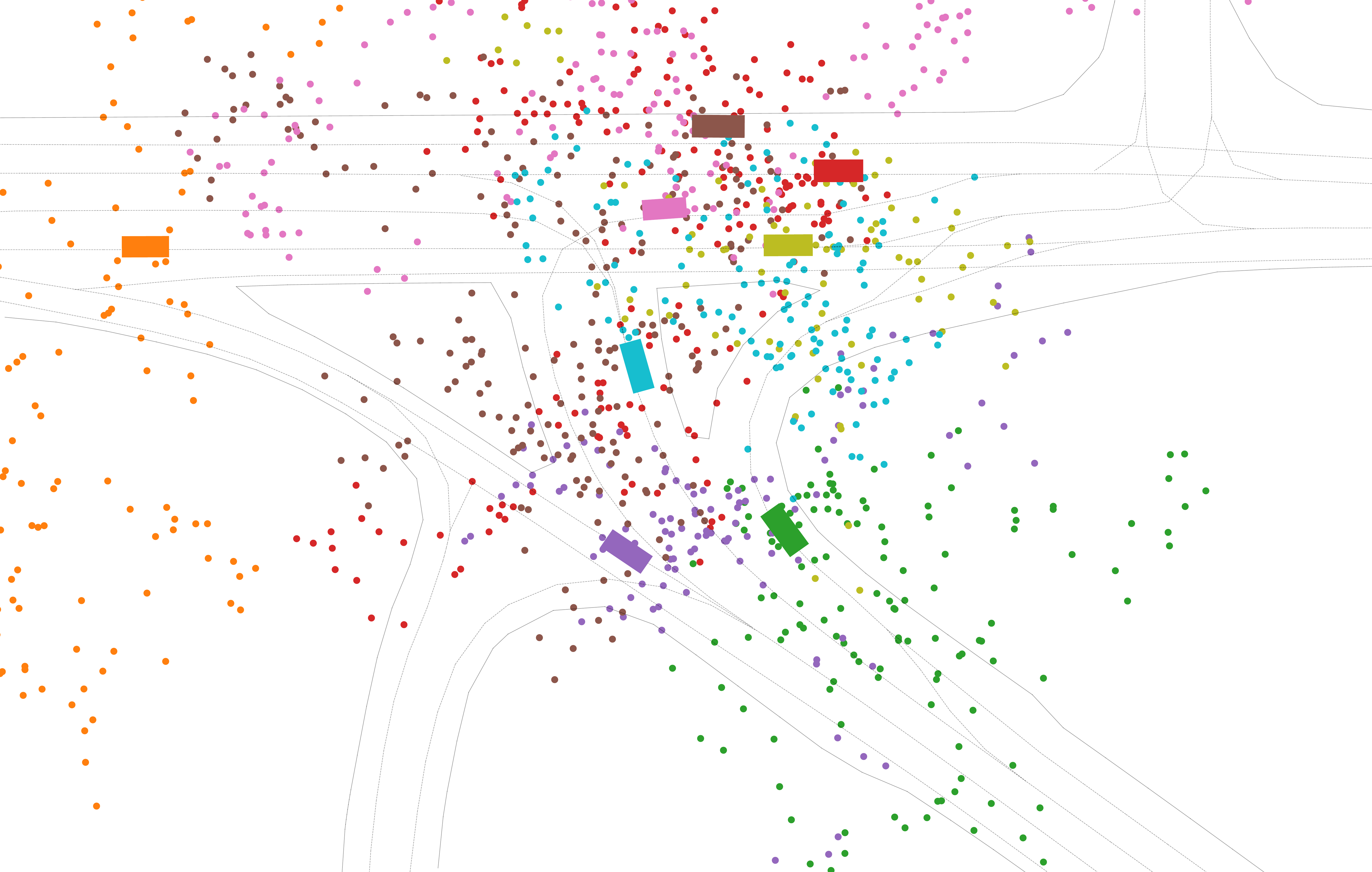}
        \caption{$k = 500$}
        \label{fig:subfig1}
    \end{subfigure}\hspace{0cm}
    \begin{subfigure}{5.85cm}
        \centering
        \includegraphics[width=\textwidth]{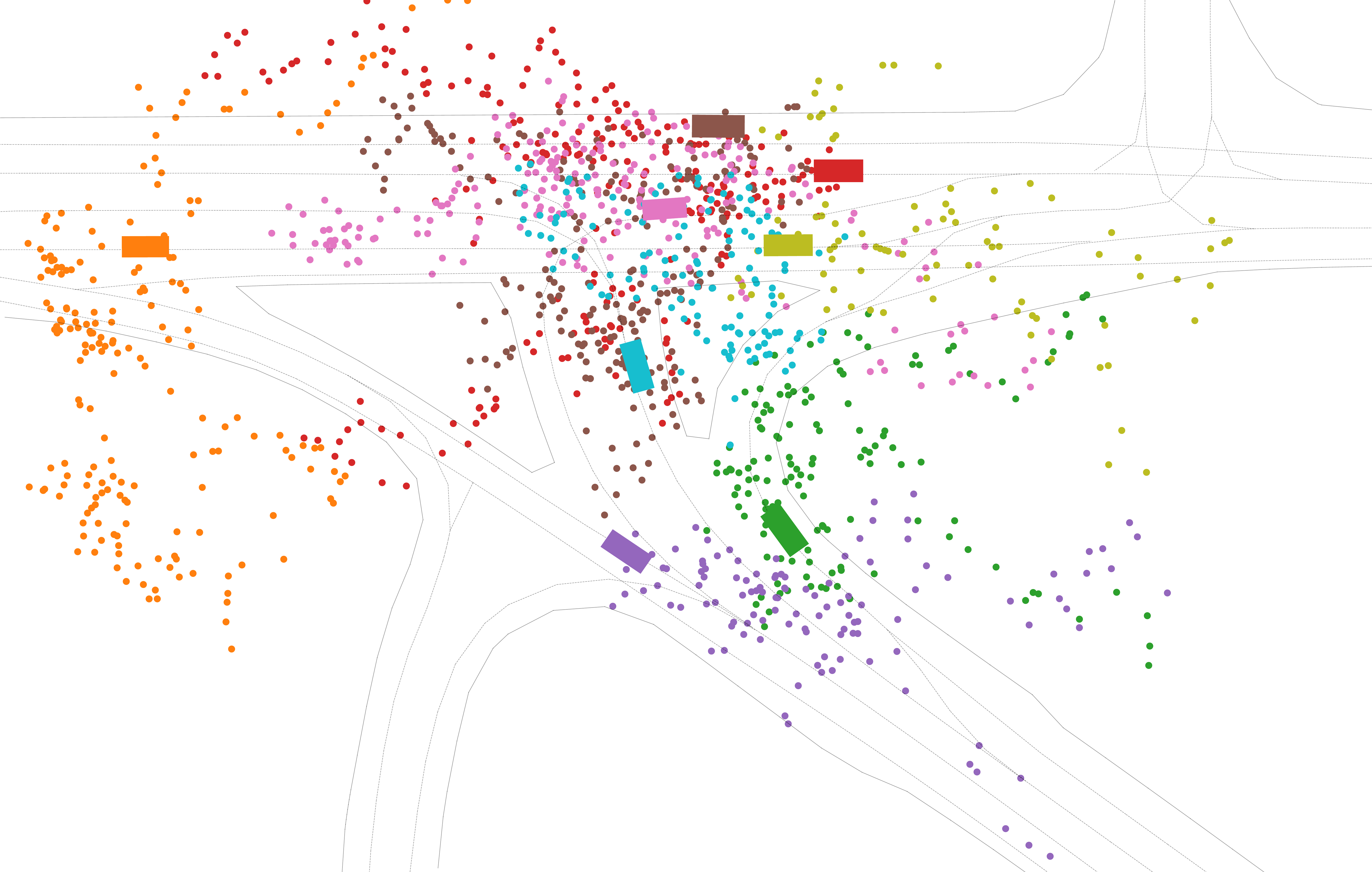}
        \caption{$k = 100$}
        \label{fig:subfig2}
    \end{subfigure}\hspace{0cm}
    \begin{subfigure}{5.85cm}
        \centering
        \includegraphics[width=\textwidth]{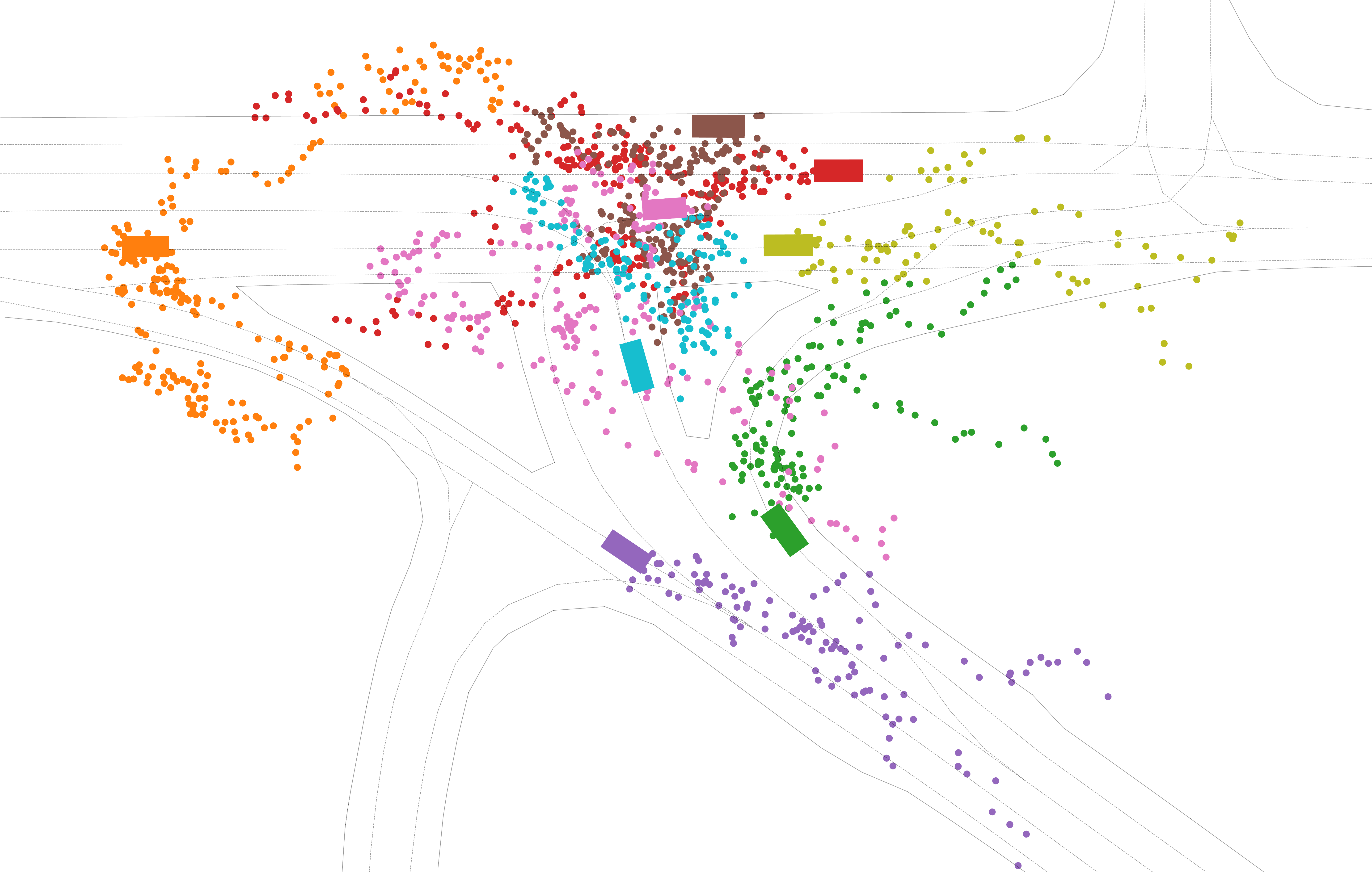}
        \caption{$k = 50$}
        \label{fig:subfig3}
    \end{subfigure}
    
    \vspace{0.01cm}
    
    \begin{subfigure}{5.85cm}
        \centering
        \includegraphics[width=\textwidth]{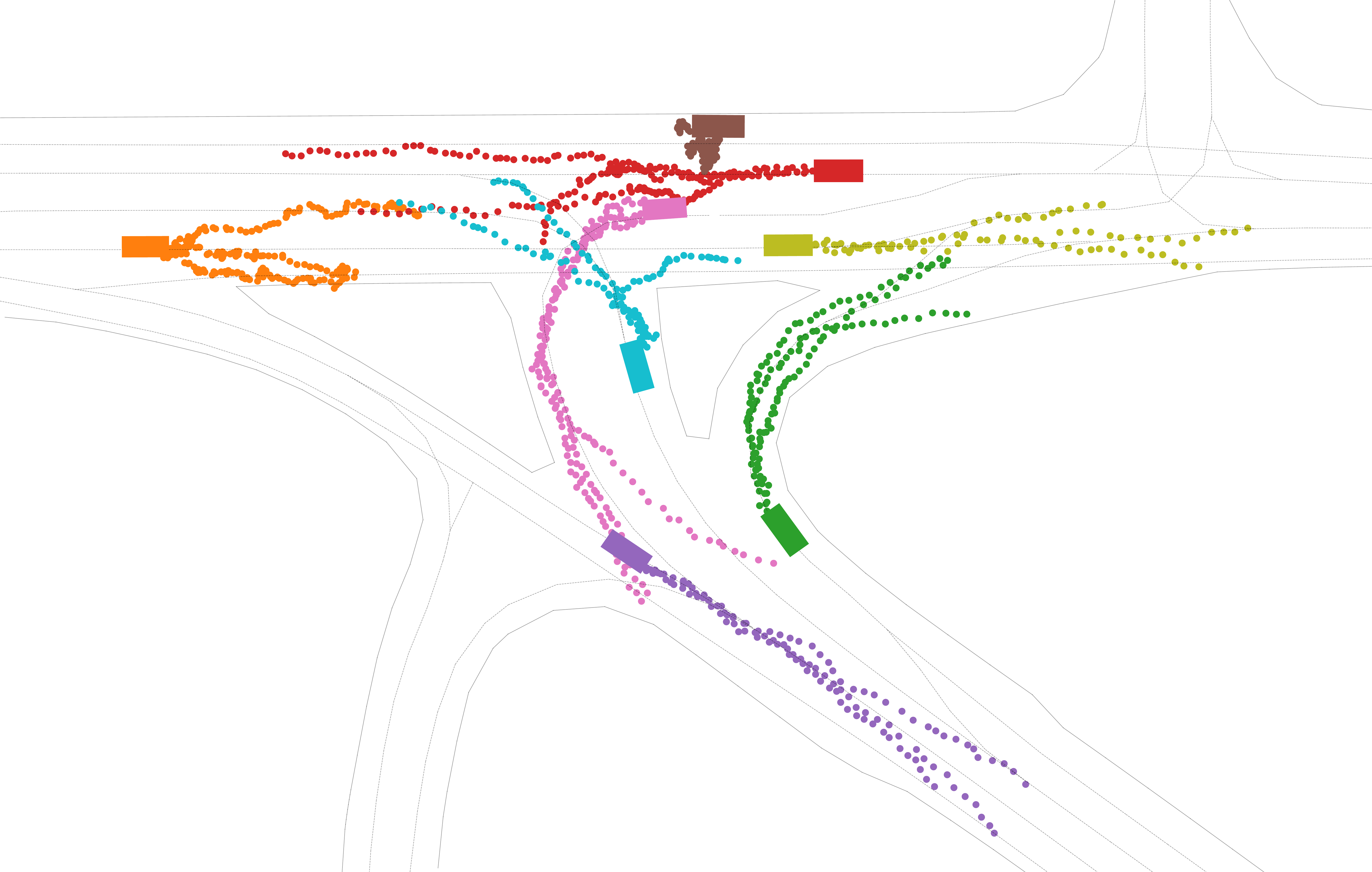}
        \caption{$k = 10$}
        \label{fig:subfig4}
    \end{subfigure}\hspace{0cm}
    \begin{subfigure}{5.85cm}
        \centering
        \includegraphics[width=\textwidth]{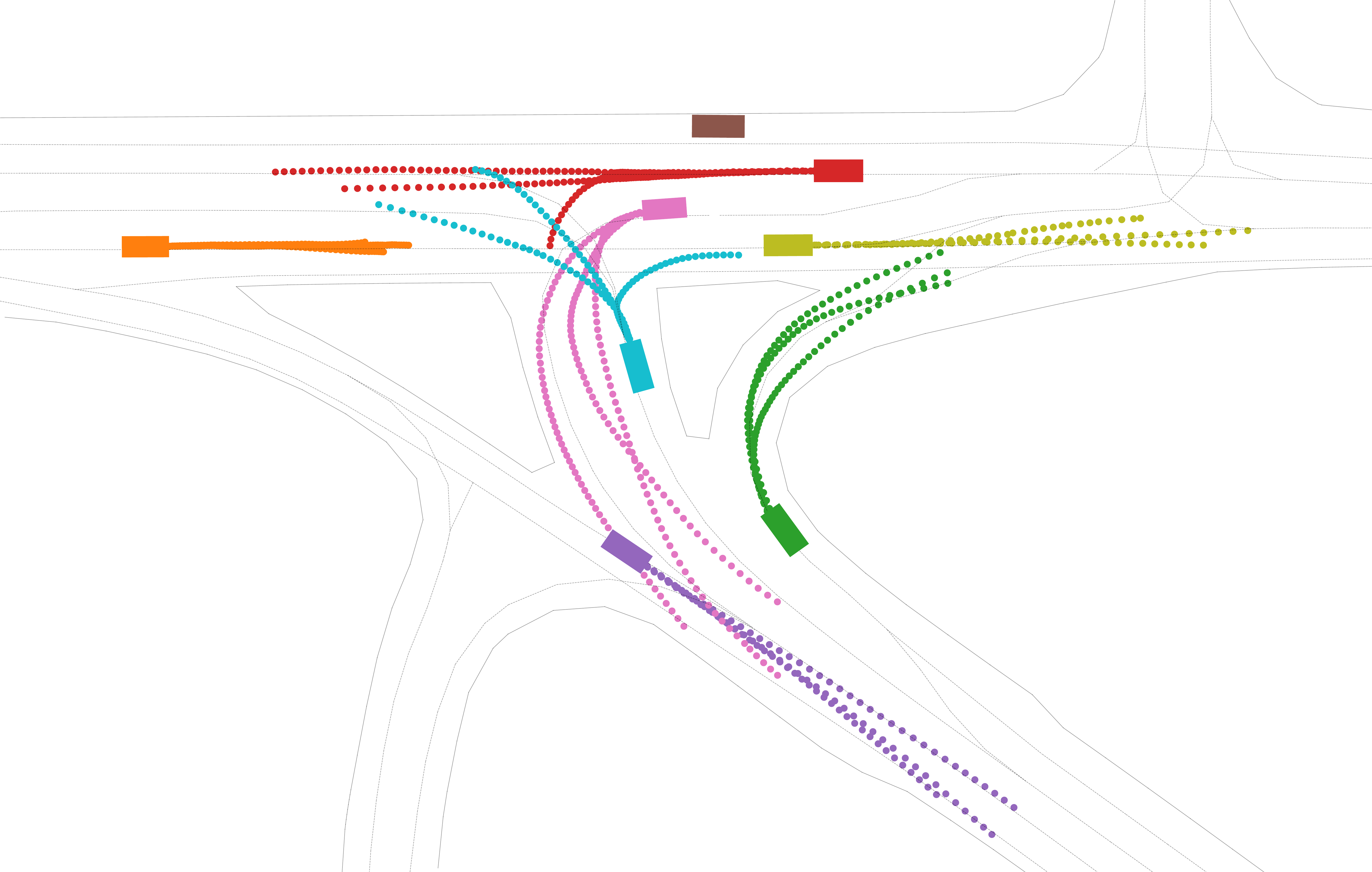}
        \caption{$k = 0$}
        \label{fig:subfig5}
    \end{subfigure}\hspace{0cm}
    \begin{subfigure}{5.85cm}
        \centering
        \includegraphics[width=\textwidth]{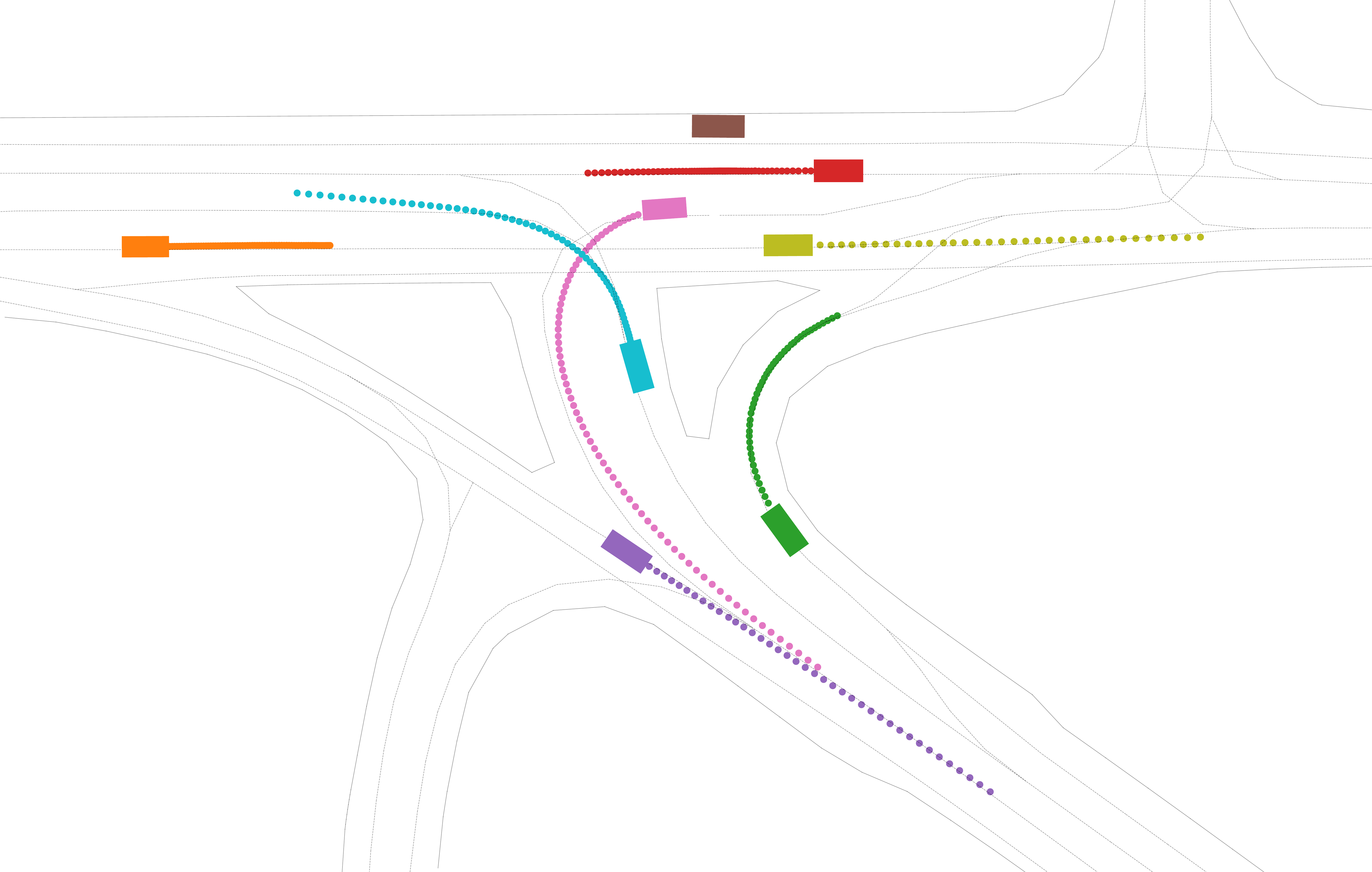}
        \caption{Ground truth}
        \label{fig:subfig6}
    \end{subfigure}

    \caption{Qualitative results of the generation process for three samples. SceneDM initially samples noise from the Gaussian distribution at $k=500$. It gradually removes trajectory noise until convergence at $k=0$.
    The comparative result with ground truth demonstrates the ability of SceneDM to capture the characteristics of real-world traffic scenes and generate realistic ones.}
    \label{fig:subfigures}
\end{figure*}

\begin{table*}[tp]
    \centering
    \caption{Ablation results of SceneDMs on 1000 scenarios that are randomly sampled from the validation set of the waymo motion dataset.}
    \label{tab:ablation}
    \setlength{\tabcolsep}{3.3mm}{
    \begin{tabular}{l|c|ccccc|cccc}
    \toprule
    Variants&\begin{tabular}[c]{@{}c@{}} Agent \\Int. \end{tabular} &
    \begin{tabular}[c]{@{}c@{}} Sequence \\Augm. \end{tabular}&
    \begin{tabular}[c]{@{}c@{}}Noise\\ const.\end{tabular} &
    \begin{tabular}[c]{@{}c@{}}Smooth\\ loss \end{tabular}  &
    \begin{tabular}[c]{@{}c@{}}Const. \\guidance \end{tabular} &
     \begin{tabular}[c]{@{}c@{}}Comp. \\filter
     \end{tabular} &
    \begin{tabular}[c]{@{}c@{}} Kinematic \\Metric \end{tabular}&
    \begin{tabular}[c]{@{}c@{}} Interactive \\Metric \end{tabular} &
    \begin{tabular}[c]{@{}c@{}} Map \\Metric \end{tabular} &
    Realism \\ 
    \midrule
    Baseline & &  &  &  &  &&0.4108  &0.5052  &0.5899 &0.4891  \\
    - &  \checkmark &   &  &  &  & &0.4159  &0.5161  &0.5964 &0.4965  \\
    -&  \checkmark &  \checkmark &  &  &  &&0.3904  &0.5047  &0.5408 &0.4682  \\
    -& \checkmark &  \checkmark &  \checkmark  &\checkmark &    &  &0.4209  &0.5181  &0.5863 &0.4963  \\
    SceneDM& \checkmark &  \checkmark &  \checkmark  & \checkmark  &  \checkmark &&0.4243  &0.5174  &0.5894 &0.4982 \\
    SceneDMF & \checkmark &  \checkmark &  \checkmark  & \checkmark  &  \checkmark & \checkmark& 0.4245 &0.5257  &0.5959 &0.5030\\
    \bottomrule
\end{tabular}}
\end{table*}

\subsection{Results and Analysis}
\textbf{Leaderboard results.} The Sim Agents Benchmark assesses all methods on the test split.
We quantitatively compare our method with a wide range of methods.
As shown in Table \ref{tab:leader_result}, the proposed models reach a \textbf{realism} meta-metric of 0.5060, achieving the state-of-the-art performance, and achieve the highest scores in most metrics. Notably, with speed as the motion state,  the proposed methods perform the best in terms of linear acceleration, which demonstrates the effectiveness of the consistent diffusion strategy in improving the smoothness of sequence data.
Interactive results indicate that SceneDM effectively handles the interaction among various types of agents. As shown in Table~\ref{tab:leader_result} and \ref{tab:ablation}, SceneDMF further improves the performance by filtering scenarios with the scoring module, especially in terms of collision and offroad metrics.

\textbf{Qualitative results.} In Figure~\ref{fig:subfigures}, we illustrate the dynamic generation process of a scenario. 
The process commences with the sampling of Gaussian noise when $k=500$.
During the denoising process, SceneDM progressively eliminates the noise and reduces the trajectory uncertainty. 
It ultimately converges when $k=0$, yielding trajectories that adhere to the distribution of real-world data.
To provide a concise representation, we randomly select three distinct trajectories from the set of 32 generated future trajectories for display. 
As demonstrated in Figure \ref{fig:subfigures},  1) when the agent goes on the straight lane, SceneDM captures diverse speed modes and is capable of lane changing;
2) On an intersection, SceneDM generates multi-mode trajectories, including  going straight, turning left, and turning right, etc.. 3) For turning trajectories, SceneDM covers diverse turning radii, consistent with those in the real world. These observations demonstrate that SceneDM effectively models the multi-modal characteristics of the agents. More qualitative results are further provided on the project webpage. 

\textbf{Analysis of Trajectory Smoothness.}
According to the kinematic metrics presented in Table \ref{tab:leader_result}, SceneDM achieves the best in acceleration metric, and the $2$-nd in terms of angular velocity and acceleration metric. This demonstrates that SceneDM is capable to generate smooth motion trajectories and aligns with the movement patterns of agents in real-world scenarios, with the proposed consistent diffusion strategy.

\subsection{Ablation Studies}
We further study the different module proposed in this paper. Due to the limited number of leaderboard submission,  we randomly sampled 1,000 scenarios from the validation set to conduct ablation experiments. As shown in Table~\ref{tab:ablation}, the baseline model that is composed of just temporal attention layers  and optimized with the DDPM strategy \cite{ho2020denoising}  performs strongly but degrades significantly compared with the proposed SceneDM. With agent-wise self-attention layers introduced into the denoiser architecture, the performance is enhanced, especially in terms of the interactive metric. This result shows that across-agent attention is beneficial for generating consistent trajectories. Furthermore, simply augmenting sequence data does not improve performance.  The proposed consistent diffusion strategy and temporal-consistent guidance sampling method effectively enhance the performance. For example, the kinematic metric is enhanced from 0.3904 to 0.4209 and then 0.4243, with consistent diffusion strategy and further temporal-consistent guidance sampling. 
% These demonstrate that maintaining temporal consistency within the noisy trajectory enhances the model's ability to learn short-term temporal dependencies and generate smooth trajectories.
Finally, with the scene-level scoring module, SceneDMF achieves increment in terms of both interactive and map-based metrics, indicating the improvement on  the traffic rule compliance of the generated scenarios.

\section{Conclusion}
In this paper, we propose diffusion based multi-agent trajectory generation framework, called SceneDM, to produce consistent future motions of all kinds of agents observed in the scene.
SceneDM handles the interactive behaviors among various types of agents through interleaved attention modules across time and agents. Moreover, we propose a simple yet effective consistent diffusion strategy with just moderate modifications to the existing diffusion model DDPM. 
It permits the model to effectively capture the temporal consistency of generated trajectories, resulting in  smooth and realistic motion patterns. Furthermore,  we propose a scene-level evaluation module that is general and compatible with other motion generation methods to enhance the traffic rule compliance of the generated scenarios, particularly in complex scenarios. Finally, 
SceneDM  achieves state-of-the-art performance on the public challenging Waymo Sim Agents Benchmark. In the future, it is worth to exploring condition generation, for example generating traffic scenarios in accordance with text descriptions.

\bibliographystyle{unsrt}
\bibliography{ref}

\end{document}